# Non-Correlated Character Recognition using Artificial Neural Network

TIRTHARAJ DASH[†]
Department of Computer Science and Engineering
Veer Surendra Sai University of Technology
Burla-768018, India
Email: tirtharajnist446@gmail.com

TANISTHA NAYAK
Department of Information Technology
National Institute of Science and Technology
Berhampur-761008, India
Email: tanisthanist213@gmail.com

*Abstract*— **This paper investigates a method of Handwritten English Character Recognition using Artificial Neural Network (ANN). This work has been done in offline Environment for non correlated characters, which do not possess any linear relationships among them. We test that whether the particular tested character belongs to a cluster or not. The implementation is carried out in Matlab environment and successfully tested. Fifty-two sets of English alphabets are used to train the ANN and test the network. The algorithms are tested with 26 capital letters and 26 small letters. The testing result showed that the proposed ANN based algorithm showed a maximum recognition rate of 85%.**

## I. INTRODUCTION

In recent year, the advancement in pattern recognition has accelerated due to many important applications which are not only challenging but also more demanding. Optical Character Recognition (OCR), Hand written Character Recognition, Data Mining, Signature Verification, Biometric Authentication are some applications of pattern Recognition which are more challenging and demanding research area. Character Recognition (CR) is becoming more and more important in the modern world. It helps human ease their jobs and solve more complex problem. Hand written digit recognition is a system widely used in the United States. The system is developed for zip code or postal code recognition that can be employed in mail sorting. This can help humans to sort with postal code that are difficult to identify. Machine simulation of human functions has become a challenging research field since the advent of digital computer. Character Recognition covers all types of machine recognition of characters in various application domains. Over the past few years, research in hand written signature has increased continuously. Generally we recognize machine printed character or handwritten character on paper document. Recognition of text in video, web document, development of electronic library, multimedia database, systems which require handwriting data entry are some newly emerging area which has become an urgent demand. Generally character recognition can be broadly classified into two types based on the data acquisition process and the text type. They are (i) online (ii) offline. In offline method, the pattern is captured as an image and taken for testing purpose. But in case of online approach, each point of the pattern is a function of pressure, time, slant, strokes and other physical parameters. Both the methods are best based on their application in the field of machine learning. Yielding best accuracy with minimal cost of time is a crucial precondition for pattern recognition system. Therefore, hand written character recognition is continuously being a broad area of research.

There are two different techniques that can be used to recognize handwritten characters. Some of them are Pattern Recognition and Artificial Neural Network techniques. Bayesian Decision theory, Nearest Neighbor rule and Linear Classification are some methods for Pattern Recognition. Face Recognition, thumb print recognition, speech recognition, Handwritten Digit Recognition uses Neural Network to recognize them.

Some of the advantages of using Neural Networks for recognition are
- It is insensible to noise.
- It can solve problem using multiple constraints.
- It can solve more complex problem.
- Accuracy rate is very high.

In this machine learning world, English Hand Written Character Recognition has been a challenging and interesting research area in the field of Artificial Intelligence and Soft computing [1,2]. It contributes majorly to the Human and Computer interaction and improves the interface between the two [3]. Other human cognition methods viz. face, speech, thumb print recognitions are also being great area of research [4, 5, 6].

This paper is organized as follows. Section 1 presented a general introduction to the character recognition systems and methods. Section 2 gives a brief survey of some methods proposed for character recognition. Section 3 describes the proposed methodology of this work. Section 4 is a result and discussion section which gives a detailed analysis of the work. The paper is concluded in section 5 with a note to future works. The literature review section describes some previous works for character recognition proposed by esteemed researchers of the world.

## II. LITERATURE SURVEY

Available literatures convey that various approaches have been made in order to accomplish the task of character recognition. Neural Network (NN) is information processing systems, which are constructed and implemented to model the human brain [7]. The computing world has a lot to gain from neural networks whose ability to learn by example which makes them more flexible and powerful. In case of Neural Network (NN), there is no need to devise the algorithm to perform a specific task i.e., no need to understand the internal mechanism of that task. Neural Network is suited for real-time systems because of their faster response and computational time. The main objective of Neural Network (NN) is to develop a computational device for modeling the human brain to perform various computational tasks at a faster rate than traditional systems. And in about all the soft computing approaches Neural Network (NN) has been a backend of character classification. This is due to its faster computation. The methods used in front end could be (a) statistical approaches (b) kernel methods (c) support methods or (d) hybrid of fuzzy logic controllers.
Artificial Neural Network (ANN) has a large number of highly interconnected processing elements called nodes or Neuron. Each neuron is connected with other by a connection link. Each connection link is associated with weights which contains information about the input signal. The information is used by the neuron net to solve the particular problem.

Multilayer Perceptron (MLP) was used for 'Bangla' alphabet recognition by [8]. MLP is a special kind of Artificial Neural Network (ANN). More specifically the MLP is a feed-forward layered network of artificial neurons. In MLP, the artificial neurons compute the sigmoid function. A MLP consists of one input layer, one output layer and a number of hidden layers. In this work, the accuracy achieved was 86.46% and 75.05% on the samples of training and testing respectively.

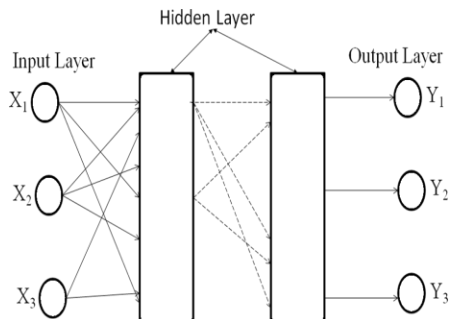

**Figure-1** A block Diagram of MLP

Neural Network based English character recognition system was proposed by [9]. In this work, MLP with one hidden layer was used. About 500 testing were carried out to test the performance of the design. The Accuracy rate obtained in this work was 94%.

English alphabet recognition using Neural Network (NN) was worked in [10]. Binary pixels of alphabets were used to train NN and the Accuracy Rate was 82.5% in this work.

A novel Chinese character recognition algorithm which was based on minimum distance classifier was proposed by Yanhua and Chuanjun [11]. The algorithm attempted to work with two classes of feature extraction- structure and statistics. The statistic feature decided the primary class and the structure feature used to identify Chinese characters.

For Handwritten numerals [12] used horizontal and vertical strokes and end points as feature. In this method, the Accuracy rate was reported as the accuracy rate of 90.5% in best case. However, this method used a thinning method resulting in loss of features.

A distribution based algorithm based on image segmentation and distribution of pixels was proposed by Huiqin [13]. Deflection Correction method was adopted for flexibility as well as reduction of matching error. This work avoided the burden of extracting the skeleton from the character. The method gave excellent result and was robust.

Manivanna and Neil proposed and demonstrated optical correlated- neural network architecture for pattern recognition [14]. English alphabet used as patterns for the training and testing process. Character recognition using Associative Memory Net was proposed by Dash [15]. The method was also applicable to parallel processing environment to make the recognition process time efficient. But, the recognition process was made offline. The highest accuracy was approximately 88.5%. Other neural network approaches such as Adaptive Resonance Theory has been applied for signature verification by [16,17,18].

A detailed analysis of some methods is given in Table-1 below which shows the References, approach and its corresponding accuracy.

**TABLE-1** Some Approaches with their Performance

| Approach | Reference | Rate of Recognition (%) |
|---|---|---|
| Multilayer Perceptron (MLP) | [8] | 75.05% |
| MLP with NN | [9] | 94% |
| Stroke Method | [10] | 90.5% |
| Neural Network with Statistical Approach | [12] | 82.5% |
| Associative Memory Net | [15] | 88.5% |

## III. PROPOSED APPROACH

We have proposed an algorithm using ANN for character Recognition i.e., non-correlated handwritten alphabets. However, our algorithm is divided into two parts
   (i) Training and
   (ii) Testing.

For training, the input is linearly dependant to output. Initially input and output are set to Zero. The network is trained with Fifty-two set of alphabets. After successful completion of Training, Testing part is done.

Any biometry or character recognition system has following principal blocks and steps.
1. Cropping the pattern
2. Feature Extraction (Pixels for character recognition)
3. Training of Network (if using NN)
4. Testing using test patterns
5. Result block

*1. Cropping the pattern*

Cropping of the pattern refers to the process of extracting the region of interest for post processing methods like feature extraction. This is carried out to avoid extra overhead during processing and unnecessary computation of white pixels. The basic aim of cropping is to get the smallest possible region of interest.

*2. Feature extraction*

The Feature Extraction of the pattern can be demonstrated as in the Figure-2 below. The conditional statement showing this extraction process is demonstrated below.

*if (colored pixel)*
*then extract '1'*
*else*
*then extract '0'*

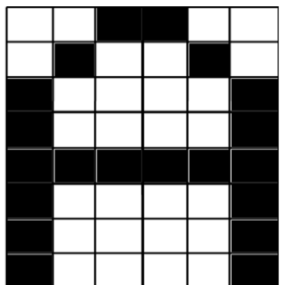

**Figure-2(a)** Character Pattern **(b)** Extracted Pixels

These pixel values were then stored in a text file for future processing by the neural network. Basically, these values will be used as input the neural network. The neural network contains number of nodes equals to size of the two dimensional pixel matrix so obtained [16,17].

A character 'A' can be written by different persons in different styles. This style also affects region of interest to be cropped and further computation. So, it will be very much crucial to make a note on different styles of writing a character by different persons. Some of the styles are shown in Figure-3 below. This is also called as textual orientation of character pattern. However, the process of feature extraction is the same for any kind of pattern.

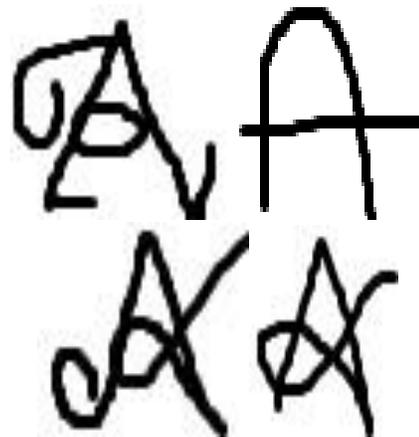

**Figure-3** Different styles of writing the character 'A'

Some characters in computerized form are given below from which the pixels could be extracted.

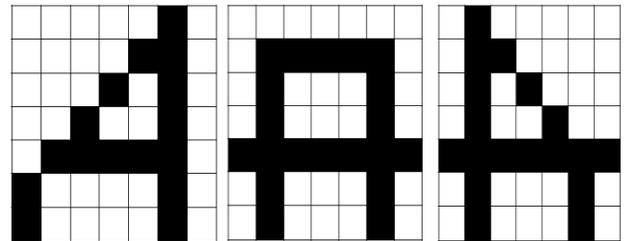

**Figure-4** Representing 'A' in different styles

*3. Algorithm:*

*Step 0:* **START**
*Step 1:* Collect the Training English Characters (52 alphabets)
*Step 2:* Collect the Test Patterns
*Step 3:* Extract the features from patterns collected in Step-1,2.
*Step 4:* Developing of the ANN algorithm for Training and Testing.
*Step 5:* Train the developed net with training patterns and store the weight (knowledge base).
*Step 6:* Perform the testing for the testing patterns and store the weights.
*Step 7:* Check the updated knowledge base with all the previous knowledge bases.
*Step 8:* **STOP**

The updation of knowledge is given by Equation-1 below.

$$K(new) = K(old) + Input \times Target \quad (1)$$

The method proposed in Algorithm above is demonstrated in Figure-3 below. The diagram shows number of training and testing patterns.

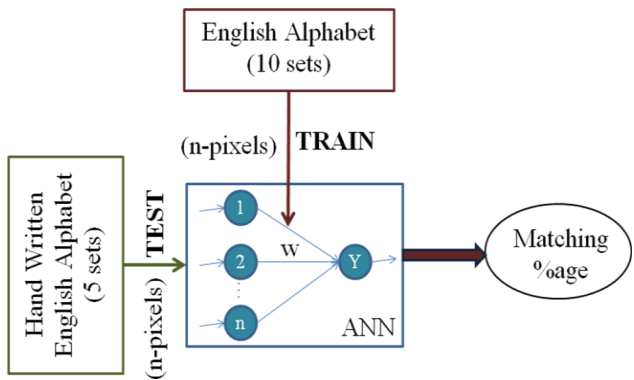

**Figure-5** Proposed Methodology

The Flowchart Diagram of the proposed algorithm is given below.

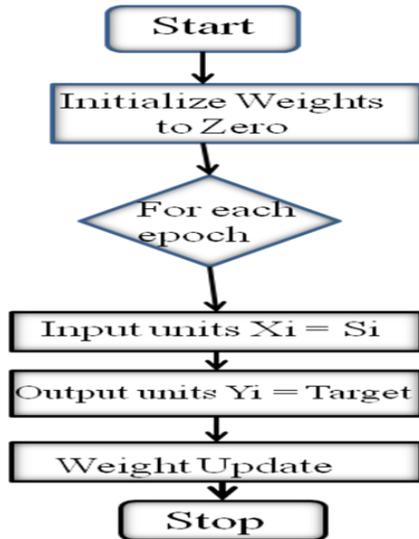

**Figure-6** Flowchart of the training process [7]

## IV. RESULT AND DISCUSSION

The proposed method has been implemented in Matlab 2010a environment. As it is already mentioned that the neural network is trained with English characters and tested with hand written English character patterns.

The network has been trained with the training patterns till the minimal mean square error is achieved. The testing of the network with the test pattern resulted with the following observations. The observations are drawn from the recognition rate achieved, which is given in Table-2 below.

**TABLE-2** Character Recognition Rate achieved by the NN

| Character | Recognition Rate (%) | False Matching with other Character |
|---|---|---|
| I | 78% | **I**, L, P, F |
| L | 82% | I, **L**, E |
| T | 85% | **T**, I, F |
| O | 67% | **O**, Q |

In the above table, only the characters which were recognized with more than 50% recognition rate are given.

The authors are doing hard work to make the neural network recognize any character belonging to ASCII character set.

## V. CONCLUSION

This work proposed an algorithm for hand written English alphabet pattern recognition. The algorithm is based on principle of Artificial Neural Network (ANN). However, the algorithm test result showed that only a few character patterns were recognized with more than 50% accuracy. If this issue is suppressed then the developed algorithm showed a maximum recognition rate of 85% for the character 'T' which is followed by 82% for the character pattern 'L'.

As future work, it will be interesting to make the algorithm that much efficient to recognize the character patterns belonging to ASCII character set.

## AUTHOR'S BIOGRAPHY

**Mr. Tirtharaj Dash** is currently an M.Tech Scholar in the Department of Computer Science and Engineering at Veer Surendra Sai University of Technology (Formerly UCE), Burla, India. He completed his B.Tech (IT) from National Institute of Science and Technology, Berhampur in the year 2012. He has contributed around 16 research and review papers in Conferences and reputed journals. His areas of research include Pattern Recognition, Algorithms, Soft Computing, Parallel Processing and Quantum Computation.

**Ms. Tanistha Nayak** graduated from National Institute of Science and Technology with a Bachelor Degree in Technology (B.Tech-IT) in the year 2012. Her research interests are based on the techniques of Soft Computing approaches and Quantum Computing. She has already published around 15 research and review papers on these fields in Conferences and journals.